\documentclass[conference]{IEEEtran}
\IEEEoverridecommandlockouts

\usepackage{indentfirst}
\usepackage{cite}
\usepackage{amsmath,amssymb,amsfonts}
\usepackage{algorithmic}
\usepackage{graphicx}
\usepackage{subfigure}
\usepackage{textcomp}
\usepackage{xcolor}
\usepackage{float}
\usepackage{enumitem}
\usepackage{multirow}
\usepackage{xcolor}

\newcommand{\Rmnum}[1]{\uppercase\expandafter{\romannumeral #1}}

\def\BibTeX{{\rm B\kern-.05em{\sc i\kern-.025em b}\kern-.08em
    T\kern-.1667em\lower.7ex\hbox{E}\kern-.125emX}}

\DeclareRobustCommand*{\IEEEauthorrefmark}[1]{%
	\raisebox{0pt}[0pt][0pt]{\textsuperscript{\footnotesize #1}}%
}

\begin{document}

\title{MTPareto: A MultiModal Targeted Pareto Framework for Fake News Detection \\
\thanks{This work is supported by National Natural Science Foundation of China (NSFC) (No.62101553, No.62306316, No.U21B20210, No.62201571). This work is also partly supported by the Major Program of the National Social Science Fund of China (13\&ZD189)}
}


\author{
    \IEEEauthorblockN{Kaiying Yan\IEEEauthorrefmark{1}, 
    Moyang Liu\IEEEauthorrefmark{2}, Yukun Liu\IEEEauthorrefmark{3*}\thanks{*corresponding author.}, 
    Ruibo Fu\IEEEauthorrefmark{4},  Zhengqi Wen\IEEEauthorrefmark{5}, 
    Jianhua Tao\IEEEauthorrefmark{5,6}, Xuefei Liu\IEEEauthorrefmark{4}, 
    Guanjun Li\IEEEauthorrefmark{4}}
    \IEEEauthorblockA{\IEEEauthorrefmark{1} School of Mathematics, Sun Yat-sen University, Guangzhou, China}
    \IEEEauthorblockA{\IEEEauthorrefmark{2} Beihang University, Beijing, China}
    \IEEEauthorblockA{\IEEEauthorrefmark{3} University of Chinese Academy of Sciences, Beijing, China}
    \IEEEauthorblockA{\IEEEauthorrefmark{4} Institute of Automation, Chinese Academy of Sciences, Beijing, China}
    \IEEEauthorblockA{\IEEEauthorrefmark{5} Beijing National Research Center for Information Science and Technology, Tsinghua University, Beijing, China}
    \IEEEauthorblockA{\IEEEauthorrefmark{6} Department of Automation,Tsinghua University, Beijing, China}
    \IEEEauthorblockA{yanky6@mail2.sysu.edu.cn, moyang\_liu@buaa.edu.cn,  yukunliu927@gmail.com}
}

\maketitle

\begin{abstract}
Multimodal fake news detection is essential for maintaining the authenticity of Internet multimedia information. Significant differences in form and content of multimodal information lead to intensified optimization conflicts, hindering effective model training as well as reducing the effectiveness of existing fusion methods for bimodal. To address this problem, we propose the MTPareto framework to optimize multimodal fusion, using a Targeted Pareto(TPareto) optimization algorithm for fusion-level-specific objective learning with a certain focus. Based on the designed hierarchical fusion network, the algorithm defines three fusion levels with corresponding losses and implements all-modal-oriented Pareto gradient integration for each. This approach accomplishes superior multimodal fusion by utilizing the information obtained from intermediate fusion to provide positive effects to the entire process. Experiment results on FakeSV and FVC datasets show that the proposed framework outperforms baselines and the TPareto optimization algorithm achieves 2.40\% and 1.89\% accuracy improvement respectively.

\end{abstract}

\begin{IEEEkeywords}
multimodal fake news detection, multimodal fusion, Pareto optimal
\end{IEEEkeywords}

\section{Introduction}
 
The rapid growth of the Internet and multimedia platforms like YouTube and TikTok \cite{jordan2024rise} has led to the rampant spread of multimedia fake news, making it harder to discern truth from falsehood \cite{wittenberg2021minimal}. To address this challenge, automated fake news detection models can enhance screening efficiency and help maintain a truthful online environment.

Early works are mostly unimodal, extracting statistical features from texts \cite{shu2019beyond,serrano2020nlp} or modelling temporal consistency \cite{yang2019exposing} for classification.
Then bimodal fusion methods emerged, like combining textual and visual features \cite{khattar2019mvae,qian2021hierarchical,singhal2020spotfake+}, identifying semantic inconsistencies\cite{xue2021detecting}. 
Direct bimodal concentration can be rigid, so many architectures are designed to capture inter- and intra-modal information \cite{singhal2022leveraging, wei2023modeling} and methods like entity-centric interaction \cite{li2021entity}, contrastive learning \cite{wang2022misinformation,wang2023cross} have been developed.
With the advancement of deep learning, fake news detectors now incorporate more modalities, like audio, video and comments, requiring improved alignment and fusion strategies between them. Accordingly, fusion networks such as progressive co-attention\cite{shang2021multimodal, you2022video, liu2024transformer}, cross-modal Transformer \cite{qi2023fakesv} and methods like mitigating multimodal biases \cite{zeng2024mitigating}, exploiting large language models\cite{hu2024bad,jin2024fake} are introduced. 

In multimodal fake news detection, conflicts also occur in model optimization due to differences in form and content between modalities. Most existing methods typically implement bimodal fusion and alignment \cite{qian2021hierarchical,wang2022misinformation}, apply bimodal fusion techniques directly to multiple modalities \cite{you2022video,liu2024transformer}, or rely on specific pipelines \cite{wei2023modeling, zeng2024mitigating}. These methods can achieve certain effectiveness since the optimization conflict is not serious in bimodal. However, since real-world media often includes text, audio, images, and other modalities simultaneously, there is an urgent need for methods that improve the efficiency of multimodal fusion.

Pareto optimization\cite{lin2019pareto,sener2018multi} resolves conflicting objectives by identifying a Pareto optimal solution between multiple objective functions, widely applied in multi-task learning\cite{ma2020efficient,dimitriadis2023pareto,swamy2024pembot}. Combining Pareto optimization with multimodal learning is a promising direction and some theoretical analysis\cite{wei2024mmpareto} have verified its effectiveness in bi-modal fusion.

In this paper, to address the multimodal parameter optimization challenge, we introduce an innovative approach to optimizing multimodal fusion through Pareto optimal training strategies and propose the \textbf{M}ultiModal \textbf{T}argeted \textbf{Pareto} (MTPareto) framework. 
Adhere to the overall purpose of optimizing all-modal fusion, we leverage the TPareto optimization algorithm to enhance feature fusion by defining fusion levels which are analogous to multiple tasks and performing all-modal-oriented Pareto gradient integration during the hierarchical fusion process. 
This method ensures that key information is progressively extracted as more modalities are introduced, mitigating the adverse effects of multimodal interactions. 
Experimental results demonstrate that TPareto enhances multimodal fusion by taking full advantage of intermediate fusion, leading to improved overall performance. More ablation experiments demonstrate performance improvement at each fusion level, proving the effectiveness of this method.

\begin{figure*}[htbp]
\vspace{-2.0em}
	\subfigure[Multimodal Hierarchical Fusion Architecture. Three fusion levels and losses are defined based on the hierarchical architecture, with the corresponding gradients used for TPareto integration.] 
	{
		\begin{minipage}[t]{0.49\linewidth}
			\centering          
			\includegraphics[width=1\linewidth]{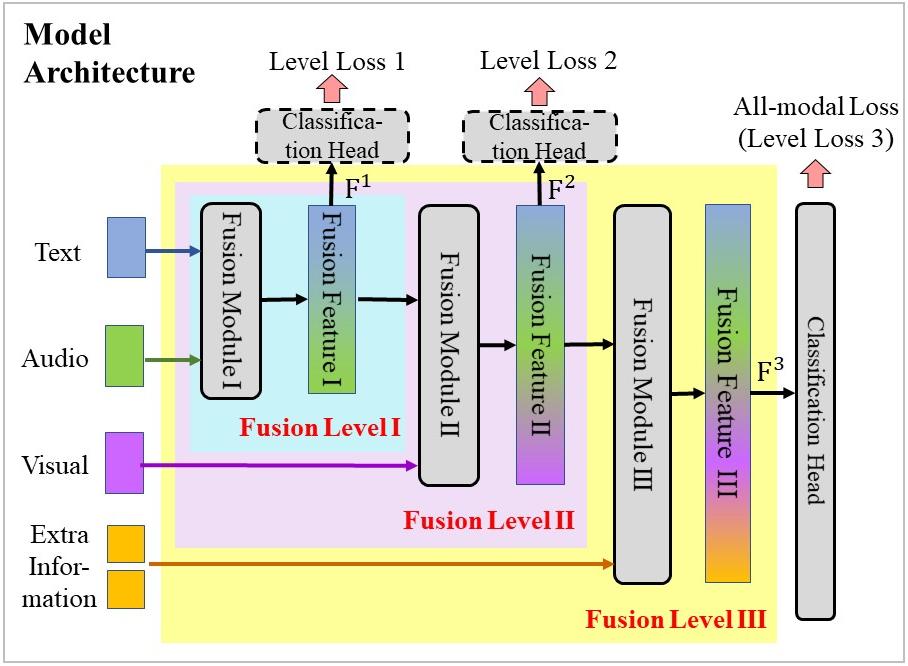}   
		\end{minipage}
            \label{fig:model}
	}
	\subfigure[TPareto Optimization Algorithm (take Fusion Module \Rmnum{1} for example). Compared with the normal additive method leading to an unexpected deviation, it is utilized with three different cases defined. In the non-conflict case, 3 level gradients are involved in the integration. In the angle-conflict case, the gradient with an angle exceeding threshold is not involved. In the weight-conflict case, the weight greater than threshold is truncated.] 
	{
		\begin{minipage}[t]{0.49\linewidth}
			\centering      
			\includegraphics[width=1\linewidth]{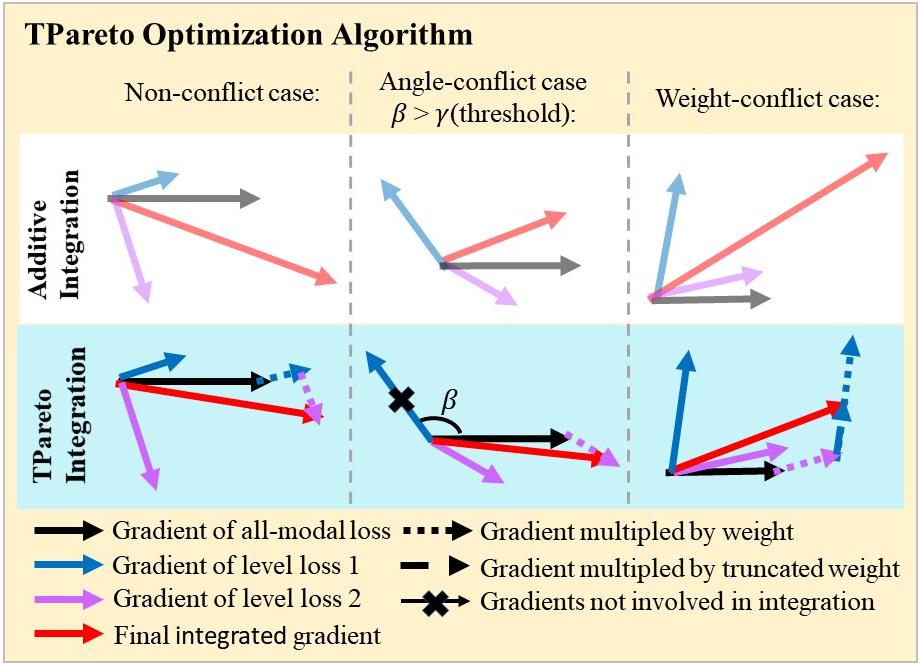}   
		\end{minipage}
            \label{fig:TPareto}
	}
    \vspace{-1.5em}
	\caption{Illustration of the MTPareto framework.} 
	\label{fig:1}  
\vspace{-2.0em}
\end{figure*}

\section{Multimodal Hierarchical Fusion Architecture}
\label{sec:network}
\subsection{Overview of Model Architecture}
\label{overview}

We use pre-trained BERT, VGG19, and Wav2Vec to extract text, visual, and audio feature embeddings respectively and the number of likes is used to weight comment data. As shown in Fig.~\ref{fig:model}, after passing through multiple hierarchical fusion modules, the features are fed into a Transformer-based classification head for final predictions. The model is trained using the cross-entropy loss function.

\subsection{Hierarchical Fusion Modules}

In this section, we provide a detailed introduction of fusion modules designed based on multi-head attention \cite{vaswani2017attention} and cross-attention \cite{lu2019vilbert}. Fig.~\ref{fig:model} illustrates the overall hierarchical architecture of the proposed framework. More details are explained as follows: 
\begin{enumerate}
    \item[1.] Two-stream Cross-attention Fusion Module for Text and Audio —— Fusion Module \Rmnum{1}

    Since both text and audio features have temporal characteristics, we use a cross-attention transformer to fuse them, resulting in audio-enhanced text features and text-enhanced audio features collectively forming the first level fusion feature $F^1$ through concentration.
    
    \item[2.] Adaptive Weighted Cross-attention Fusion Module for Text, Audio and Images —— Fusion Module \Rmnum{2}

    In the second layer of fusion, the image features are fused with the enhanced text and audio features, resulting in further enhanced text, audio and the two enhanced image features. The adaptive weights of the two image feature are determined through cross-attention and these weighted features are combined into the fused image feature. The three-modal fusion obtains the second level fusion feature $F^2$.

    \item[3.] Gated-Cross-Attention Fusion Module for Extra Information —— Fusion Module \Rmnum{3}

    Extra features(e.g., comments, publisher profiles) aid in fake news detection, but their importance varies depending on the video. To manage the dynamic interaction between the main modalities and these auxiliary features, we use a gating mechanism\cite{arevalo2017gated} with softmax for handling dynamic dependencies and filtering the unimportant inputs. This module ensures that auxiliary features are integrated without overshadowing the main modalities and the third level fusion feature $F^3$ is obtained after fusing text, audio, image and extra information. 
    
\end{enumerate}

\section{Targeted Pareto Optimization Algorithm}
\label{sec:method}

\subsection{Pareto Integration Preliminaries}

Pareto method \cite{sener2018multi} is utilized to achieve balance in multi-task learning by finding the optimal trade-off point between several losses and weighting the gradients of multiple tasks. The overall optimization goal is a
convex quadratic problem with linear constraints, written as follows:
\begin{align}\label{MGDA}
    \min_{\alpha^{1},...,\alpha^{T}}\left\{\left\|\sum_{t=1}^{T}\alpha^{t}\nabla_{\mathbf{\boldsymbol{\theta}_{sh}}}\hat{\mathcal{L}}^{t}(\boldsymbol{\theta}_{sh},\boldsymbol{\theta}_{t})\right\|_{2}^{2}\right|\sum_{t=1}^{T}\alpha^{t}=1,\alpha^{t}\geq0\quad\forall t\biggr\}
\end{align}
where $\theta^{sh}$ is shared parameters between tasks and $\theta^{t}$ is task-specific, $\hat{\mathcal{L}^{t}}(\boldsymbol{\theta}_{sh},\boldsymbol{\theta}_{t})$ is the loss of t-th task.

Considering two tasks, the optimization problem can be defined as:
\begin{align}
    \min_{\alpha\in[0,1]}\|\alpha\nabla_{\boldsymbol{\theta}_{sh}}\hat{\mathcal{L}}^1(\boldsymbol{\theta}_{sh},\boldsymbol{\theta}_1)+(1-\alpha)\nabla_{\boldsymbol{\theta}_{sh}}\hat{\mathcal{L}}^2(\boldsymbol{\theta}_{sh},\boldsymbol{\theta}_2)\|_2^2,
\end{align}
and an analytical solution can be obtained for this one-
dimensional quadratic function:
\begin{align}
\footnotesize
\hat{\alpha}=\left[\frac{\left(\nabla_{\boldsymbol{\theta}_{sh}}\hat{\mathcal{L}}^{2}(\boldsymbol{\theta}_{sh},\boldsymbol{\theta}_{2})-\nabla_{\boldsymbol{\theta}_{sh}}\hat{\mathcal{L}}^{1}(\boldsymbol{\theta}_{sh},\boldsymbol{\theta}_{1})\right)^{\mathsf{T}}\nabla_{\boldsymbol{\theta}_{sh}}\hat{\mathcal{L}}^{2}(\boldsymbol{\theta}_{sh},\boldsymbol{\theta}_{2})}{\|\nabla_{\boldsymbol{\theta}_{sh}}\hat{\mathcal{L}}^{1}(\boldsymbol{\theta}_{sh},\boldsymbol{\theta}_{1})-\nabla_{\boldsymbol{\theta}_{sh}}\hat{\mathcal{L}}^{2}(\boldsymbol{\theta}_{sh},\boldsymbol{\theta}_{2})\|_{2}^{2}}\right]_{+,\frac{1}{\mathbf{r}}}
\label{analytical}
\end{align}
where $[\cdot]_{+,\frac{1}{\tau}}$represents clipping to $[0,1]$. 

When the number of tasks exceeds two, the weights can be efficiently determined by applying the Frank-Wolfe algorithm \cite{jaggi2013revisiting}, utilizing \eqref{analytical} as a subroutine for the line search which can be solved analytically.

\subsection{Targeted Pareto Integration in Hierarchical-Fusion-based Multimodal Learning} 
\label{sec:TPareto}

From an innovative perspective, we analyze the hierarchical fusion network from the Pareto optimal perspective and propose \textbf{T}argeted \textbf{Pareto} (TPareto) optimization algorithm. Our method conducts gradient integration considering both the direction and relative weight of the gradient fusion process to provide all-modal learning with harmless assistance from the hierarchical fusion process. The overall TPareto optimization algorithm is shown in Fig.~\ref{fig:TPareto}. In the following section, we use the three-layer hierarchical fusion network we designed as an example for analysis.

\textbf{Multiple Levels and Loss Functions for Hierarchical Fusion Networks}: Based on the definition in Sec \ref{sec:network}, we let ${\theta}_1,{\theta}_2,{\theta}_3,{\theta}_{cls}$ represent the parameters of fusion module \Rmnum{1}, \Rmnum{2}, \Rmnum{3} and classfication head respective. As shown in Fig.~\ref{fig:TPareto}, different losses are defined corresponding to different fusion levels \Rmnum{1}, \Rmnum{2}, \Rmnum{3}, donated as  $\hat{\mathcal{L}}^{1}(\boldsymbol{\theta}_{1},\boldsymbol{\theta}_{cls}) $, $\hat{\mathcal{L}}^{2}(\{\boldsymbol{\theta}_{i}\}_{1}^{2},\boldsymbol{\theta}_{cls})$,  $\hat{\mathcal{L}}^{All}(\{\boldsymbol{\theta}_{i}\}_{1}^{3},\boldsymbol{\theta}_{cls})=\hat{\mathcal{L}}^{3}(\{\boldsymbol{\theta}_{i}\}_{1}^{3},\boldsymbol{\theta}_{cls})$.

\textbf{Targeted Pareto Integration of Fusion Modules}: In our case, the goal of applying Targeted Pareto integration is to optimize all-modal fusion while enabling each fusion module to obtain more useful feature information. 
The level losses are obtained by classifying using the fusion features of each level and different fusion module has different training objectives since it is related to different level losses. Note that fusion module \Rmnum{1} parameters ${\theta}_1$ is the shared parameters of loss $\hat{\mathcal{L}}^{1},\hat{\mathcal{L}}^{2},\hat{\mathcal{L}}^{All}$, and fusion module \Rmnum{2} parameters ${\theta}_2$ is the shared parameters of loss $\hat{\mathcal{L}}^{2},\hat{\mathcal{L}}^{All}$. The optimization goals refer to \eqref{MGDA} and are written as follows:

\begin{itemize}[leftmargin=*]
\item Fusion module \Rmnum{1}:
\begin{align}
\label{opt1}
\tiny
     \min_{\alpha^1_1,\alpha^2_1,\alpha^3_1}\left\{\left\| \alpha^1_1 \nabla_{\mathbf{\boldsymbol{\theta}_{1}}}\hat{\mathcal{L}}^{1} + \alpha^2_1 \nabla_{\mathbf{\boldsymbol{\theta}_{1}}}\hat{\mathcal{L}}^{2} + \alpha^3_1 \nabla_{\mathbf{\boldsymbol{\theta}_{1}}}\hat{\mathcal{L}}^{All} \right\|_2^2\right|\sum_{t=1}^{T_1}\alpha^t_1=1,\alpha^t_1\geq0\quad\forall t\Big\}
\end{align}

\item Fusion module \Rmnum{2}:
\begin{align}
\label{opt2}
\small
     \min_{\alpha^2_2,\alpha^3_2}\left\{\left\|\alpha^2_2\nabla_{\mathbf{\boldsymbol{\theta}_{2}}} \hat{\mathcal{L}}^{2} + \alpha^3_2 \nabla_{\mathbf{\boldsymbol{\theta}_{2}}} \hat{\mathcal{L}}^{All} \right\|_2^2\right|\sum_{t=1}^{T_2}\alpha^t_2=1,\alpha^t_2\geq0\quad\forall t\Big\}
\end{align}

\item Fusion module \Rmnum{3}:

Normal stochastic gradient descent, without Pareto gradient integration, is used to optimize its parameters since it is associated with only one loss. Therefore, the weight $\alpha^3_3$ is fixed at 1.

\end{itemize}

Through level losses backwards, we obtain gradients of the fusion module related to different level losses and $g^i_j$ refers to the gradient of fusion module j associated with level loss i. And the corresponding weights of gradients are given by \eqref{opt1},\eqref{opt2}. In our scenario, the all-modal gradient is prioritized, with its weight set to 1, while other gradients are weighted relative to it as follows:

\begin{align}
    & g^{Pareto}_{1} = {\frac{\alpha^{1}_{1}}{{\alpha}^{3}_{1}}} g^{1}_{1}+ \frac{{\alpha}^{2}_{1}}{{\alpha}^{3}_{1}}  g^{2}_{1} +  g^{All}_{1} = \frac{\sum_{i=3-T_1 +1}^{3}\alpha^{i}_{1} g^i_1}{{\alpha}^{3}_{1}}
    \\& g^{Pareto}_{2} = \frac{{\alpha}^{2}_{2}}{{\alpha}^{3}_{2}} g^{2}_{2}+  g^{All}_{2} = \frac{\sum_{i=3-T_2 +1}^{3}\alpha^{i}_{2} g^i_2}{{\alpha}^{3}_{2}} 
    \\& g^{Pareto}_{3} = g^{All}_{3}= g^{3}_{3} = \frac{\sum_{i=3-T_3 +1}^{3}\alpha^{i}_{3} g^i_3}{{\alpha}^{3}_{3}}
\end{align}
where ${\alpha}^{i}_{j}$ is the weight for gradient $g^i_j$ and $T_j$ is the gradient's number of fusion module j ($T_1$=3, $T_2$=2, $T_3$=1).

Note that our ultimate goal is to achieve more effective all-modal fusion, so we must ensure that all-modal gradients are dominant with minimal site effect from non-all-modal gradients. So we impose the following restrictions on the integration of gradients:

\begin{itemize}[leftmargin=*]
    \item  Weight-conflict case. Ensure that the sum of non-all-modal gradient relative weight is less than threshold k by doing numerical truncation, donated as non-all-modal weight threshold, to prevent the model from being biased to intermediate levels.
    
    \item Angle-conflict case. Limiting the angle between the all-modal gradient $g^{All}_j$ and non-all-modal gradient $g^{i}_j, i=1,2$ . Only if the cosine value of angle ${\beta}^{i}_j$ between $g^{All}_j$ and $g^{i}_j$ is greater than $\gamma$, $g^{i}_j$ can be added to $g^{Pareto}_{j}$ with corresponding weight and $\gamma$ is called the angle cosine threshold for measuring gradient conflict. 
\end{itemize}

\section{Experiment}
\label{sec:experiment}

\subsection{Datasets and Evaluation}

\subsubsection{Datasets.}

We do experiments on two fake news video dataset FakeSV \cite{qi2023fakesv} and FVC \cite{papadopoulou2019corpus}:
\begin{itemize}[leftmargin=*]
    \item \textbf{FakeSV}: FakeSV is the Chinese fake news short video dataset, containing 1,827 fake, 1,827 real, and 1,884 debunked videos, including multimodal information such as audio, text, images, comments, and user profiles.    
    
    \item \textbf{FVC}: FVC dataset comprises multi-lingual videos from three platforms (YouTube, Facebook, and Twitter) with textual news content and
    user comments attached, totally containing 3,957 fake and 2,458 real videos among which there are only 200 unique fake videos and 180 real unique videos.
\end{itemize}

\begin{table*}[h] 
\vspace{-1.0em}
\centering
\caption{Experimental results of baselines and the proposed MTPareto on FakeSV and FVC datasets.}
\vspace{-1.0em}
\scalebox{1.2}{
\begin{tabular}{c|cccc|cccc}
\hline
Dataset$\rightarrow$      & \multicolumn{4}{c}{FakeSV}         & \multicolumn{4}{|c}{FVC}             \\ \hline
Method$\downarrow$, Metric$\rightarrow$    & f1    & recall & precision & acc   & f1     & recall & precision & acc   \\ \hline
TikTec\cite{shang2021multimodal} & 75.02 & 75.04  & 75.11     & 75.04 & 77. 02 & 73.95  & 74.24     & 73.67 \\
FANVN\cite{choi2021using}        & 78.31 & 78.32  & 78.37     & 78.32 & 85.81  & 85.32  & 85.20     & 85.44 \\
SV-FEND\cite{qi2023fakesv}                                           & 81.02 & 81.05  & 81.24     & 81.05 & 84.71  & 85.37  & 84.25     & 86.53 \\
MMAD\cite{zeng2024mitigating}      & 82.63     & 82.73      & 82.63         & 82.64     & 90.36  & 90.46  & 90.27     & 89.28 \\ \hline
Our (w/o TPareto)             & 81.56 & 81.22  & 82.34     & 82.10 & 91.12  & 91.94  & 92.32     & 92.46 \\
\textbf{Our}       & 84.15 & 83.94  & 84.48    & 84.50 & 93.97  & 93.54  & 94.50     & 94.35 \\

\hline
\end{tabular}}
\vspace{-1.0em}
\label{tab:baselines}
\end{table*}

\subsubsection{Experimental Setup.}

For the FakeSV dataset, we split it into training, validation, and test sets in chronological order with a ratio of 70\%:15\%:15\%. For the much smaller FVC dataset, we use five-fold cross-validation and exclude unavailable user profile features. We use Bert-base-Chinese for FakeSV and Bert-base-uncased for FVC to obtain text embeddings of different languages. The model is optimized using the Adam optimizer with a learning rate of 0.0001, weight decay of 5e-3, and batch size of 64. The hyperparameters angle cosine threshold and relative weight threshold of the TPareto are set to 0.25 and 1 respectively. Evaluation is conducted using F1-score, recall, precision, and accuracy.

\subsection{Results}

\subsubsection{Comparison with Baselines}
We selected the multimodal methods for fake video detection, TikTec \cite{shang2021multimodal}, FANVN \cite{choi2021using}, SV-FEND \cite{qi2023fakesv} and MMAD \cite{zeng2024mitigating} as baselines for comparison experiments. Tab.~\ref{tab:baselines} shows that our proposed TPareto optimization algorithm achieved a 2.40\% and 1.89\% improvement in accuracy over the hierarchical fusion model and the whole framework outperforms the corresponding baseline SV-FEND \cite{qi2023fakesv} by 3.45\% and 7.82\% on FakeSV and FVC datasets respectively. Although our network(w/o TPareto) is slightly inferior to MMAD \cite{zeng2024mitigating}, it surpasses in performance when using the TPareto optimization algorithm.

\subsubsection{Ablation Study}

We further study the changes in the hierarchical fusion levels related to the TPareto algorithm, through training and evaluating different fusion levels separately. The results in Tab.~\ref{tab:Ablation} show that models with more modalities and more fusion modules may perform worse, which means there is indeed a multimodal optimization problem.

Using the TPareto, our framework outperforms baselines and improves its capabilities as more modalities are progressively integrated, referring to the results recorded in Tab.~\ref{tab:Ablation}. It is worth noting that not only the performance of the final all-modal fusion is optimized, but Fusion Level \Rmnum{1} and Fusion Level \Rmnum{2} also achieve higher accuracy. Therefore, we conclude that the TPareto optimization can avoid certain gradient conflicts and conduct purposeful integration from hierarchical fusion levels, enabling each fusion module to better capture interactions of input modalities and contribute meaningfully to the entire fusion process.

\begin{table}[H] 
\centering
\vspace{-1.0em}
\caption{Performance Comparison of different fusion levels between standard optimization and TPareto on FakeSV dataset (T:text,A:audio,V:vision,E:extra)}
\vspace{-1.0em}
\scalebox{0.97}{
\begin{tabular}{cccccc}
\hline
Fusion Level  & Method    & f1    & recall & precision & acc   \\ \hline
\multirow{3}{*}{\begin{tabular}[c]{@{}c@{}}\Rmnum{1}\\ (T+A)\end{tabular}}    & SV-FEND   & 80.32 & 80.64  & 81.02     & 81.16 \\
                      & Our (w/o TPareto)        & 80.62 & 80.30  & 81.34     & 81.18 \\
                      & \textbf{Our} & 82.43 & 82.20  & 82.83     & 82.84 \\ \hline
\multirow{3}{*}{\begin{tabular}[c]{@{}c@{}}\Rmnum{2}\\ (T+A+V)\end{tabular} }   & SV-FEND   & 79.42 & 79.87  & 79.43     & 79.53 \\
                      & Our (w/o TPareto)        & 81.85 & 81.61  & 82.29     & 82.29 \\
                      & \textbf{Our} & 83.72 & 83.43  & 84.25     & 84.23 \\ \hline
\multirow{3}{*}{\begin{tabular}[c]{@{}c@{}}\Rmnum{3}\\ (T+A+V+E)\end{tabular}}  & SV-FEND   & 81.02 & 81.05  & 81.24     & 81.05 \\
                     & Our (w/o TPareto)        & 81.56 & 81.22  & 82.34     & 82.10 \\
                       & \textbf{Our} & 84.15 & 83.94  & 84.48     & 84.50 \\ \hline
\end{tabular}}
\vspace{-1.0em}
\label{tab:Ablation}
\end{table}

\subsection{Exploration Study}

\subsubsection{Effect of the angle cosine threshold for measuring gradient conflict}
The angle cosine threshold is a critical hyperparameter that determines gradient conflict, preventing becoming biased towards intermediate non-all-modal levels. 
We can tell from Tab.~\ref{tab:cosine abortion} that an angle cosine threshold of 0.25 appears to be near-optimal, as it achieves level-by-level improvement and optimal all-modal performance. The smaller threshold may cause the model to tend to optimize non-all-modal levels, while the bigger one imposes too stringent restrictions and thus could not achieve effective Pareto gradient integration.

\begin{table}[H]
\centering
\vspace{-1.0em}
\caption{The Acc.(\%) w.r.t. angle cosine threshold used in TPareto on the FakeSV dataset (T:text,A:audio,V:vision,E:extra)}
\vspace{-1.0em}
\scalebox{1.2}{
\begin{tabular}{ccccc}
\hline
\begin{tabular}[c]{@{}c@{}}cosine threshold$\rightarrow$\\ Fusion Level$\downarrow$\end{tabular} & -0.25 & 0  & 0.25  & 0.5   \\ \hline
\Rmnum{1} (T+A)  & 82.84 	& 83.31 	& 82.84 	& 82.65  \\
\Rmnum{2} (T+A+V)   & 84.13 	& 83.57 	& 84.23 	& 83.94 \\
\Rmnum{3} (T+A+V+E)  & 83.76 	& 83.94 	& 84.50 	& 83.57 \\ \hline
\end{tabular}}
\vspace{-1.0em}
\label{tab:cosine abortion}
\end{table}

\subsubsection{Effect of non-all-modal relative weight threshold}

The important role of the relative weight restriction is to maintain the dominant role of all-modal fusion, preventing the gradient of non-all-modal levels from overtaking the all-modal level for parameter optimization. Tab.~\ref{tab:weight abortion} shows that proper weight is vital for the TPareto, as too large a value will cause a deviation from the optimization target, while too small a value will limit the performance of the TPareto algorithm.

\begin{table}[H]
\centering
\vspace{-1.0em}
\caption{The Acc.(\%) w.r.t. non-all-modal weight threshold used in TPareto on the FakeSV dataset (T:text,A:audio,V:vision,E:extra)}
\vspace{-1.0em}
\scalebox{1.2}{
\begin{tabular}{cccccc}
\hline
\begin{tabular}[c]{@{}c@{}}weight threshold$\rightarrow$\\ Fusion Level$\downarrow$\end{tabular} & 0.5   & 1  & 1.5  & 2   \\ \hline
\Rmnum{1} (T+A)  & 82.73 	& 82.84 	& 83.58 	& 83.32   \\
\Rmnum{2} (T+A+V)  & 82.18 	& 84.23 	& 83.33 	& 84.13  \\
\Rmnum{3} (T+A+V+E)  & 82.73 	& 84.50 	& 83.76 	& 83.94  \\ \hline
\end{tabular}}
\label{tab:weight abortion}
\end{table}


\section{Conclusion}
\label{sec:conclusion}

In this study, we propose a MTPareto framework, to address multimodal optimization conflicts in fake news detection by adopting a Pareto optimal perspective. 
This framework establishes fusion levels and employs a Targeted Pareto (TPareto) optimization algorithm, to promote all-modal fusion through progressively extracting key information as modalities are hierarchically incorporated. 
Experiments demonstrate that our proposed framework outperforms baseline models and the TPareto algorithm achieves a significant improvement. Further experiments verify steady enhancement within the hierarchical fusion process, with performance gains at each fusion level. This approach holds potential for application in other multimodal scenarios in the future.

\bibliographystyle{IEEEtran}
\bibliography{refs}

\vspace{12pt}

\end{document}